\documentclass[conference]{IEEEtran}
\IEEEoverridecommandlockouts
\usepackage{cite}
\usepackage{amsmath,amssymb,amsfonts}
\usepackage{algorithmic}
\usepackage{graphicx}
\usepackage{textcomp}
\usepackage{lmodern}
\usepackage{xcolor}

\usepackage{etoolbox}
\makeatletter
\patchcmd{\@makecaption}
  {.\qquad}
  {\par}
  {}{}

\long\def\@makecaption#1#2{%
  \ifx\@captype\@IEEEtablestring%
    \footnotesize{\centering\normalfont\footnotesize#1.\qquad\scshape #2\par}%
    \@IEEEtablecaptionsepspace
  \else
  \fi}

\def\BibTeX{{\rm B\kern-.05em{\sc i\kern-.025em b}\kern-.08em
    T\kern-.1667em\lower.7ex\hbox{E}\kern-.125emX}}

\begin{document}
\renewcommand\IEEEkeywordsname{\\ Keywords}

\title{
Detecting Chronic Kidney Disease(CKD) at the Initial Stage: A Novel Hybrid Feature-selection Method and Robust Data Preparation Pipeline for  Different ML Techniques
}
 \author{
            \IEEEauthorblockN{
            \textsuperscript{1}Md. Taufiqul Haque Khan Tusar, 
            \textsuperscript{2}Md. Touhidul Islam, 
            \textsuperscript{3}Foyjul Islam Raju
            }
             
        \IEEEauthorblockA{
                      \textit{Department of Computer Science and Engineering} \\
                    \textit{City University,}
                    Dhaka-1216, Bangladesh \\
                \textsuperscript{1}taufiq@ieee.org, 
                \textsuperscript{2}touhid.cse@cityuniversity.edu.bd, 
                \textsuperscript{3}raju.cu16@gmail.com
                }
    }

\maketitle

\makeatother

\begin{abstract}
Chronic Kidney Disease (CKD) has infected almost 800 million people around the world. Around 1.7 million people die each year because of it. Detecting CKD in the initial stage is essential for saving millions of lives. Many researchers have applied distinct Machine Learning (ML) methods to detect CKD at an early stage, but detailed studies are still missing. We present a structured and thorough method for dealing with the complexities of medical data with optimal performance. Besides, this study will assist researchers in producing clear ideas on the medical data preparation pipeline. In this paper, we applied KNN Imputation to impute missing values, Local Outlier Factor to remove outliers, SMOTE to handle data imbalance, K-stratified K-fold Cross-validation to validate the ML models, and a novel hybrid feature selection method to remove redundant features. Applied algorithms in this study are Support Vector Machine, Gaussian Naive Bayes, Decision Tree, Random Forest, Logistic Regression, K-Nearest Neighbor, Gradient Boosting, Adaptive Boosting, and Extreme Gradient Boosting. Finally, the Random Forest can detect CKD with 100\% accuracy without any data leakage.
\end{abstract}
\section{Introduction}
The term chronic kidney disease (CKD) indicates long-period renal impairment that has the potential to worsen. End-stage renal disorder (ESRD) is the condition of kidney whereby it stop working. The kidneys filter body wastes and excess water then expel them through urine. When CKD progresses, excessive amounts of fluid, electrolytes, and wastes may accumulate in the body. According to the statement of [1], CKD stands for a disorder where kidneys are no longer capable of filtering blood as efficiently as they should. Other health issues might arise because of CKD. Although CKD can affect anyone but some people are in a more danger than others. Those who suffer from diabetes, hypertension, heart disease, having kidney disease, age over 60 years, low calcium levels, high potassium levels, anemia, medication side effects, and so on. According to the latest data released by WHO in 2018, the number of deaths due to kidney disease has reached 16,948 or 2.18\% of the total deaths in Bangladesh [2]. Besides, CKD raises the risk of cardiovascular disease. Also, in certain circumstances, it leads to dialysis or transplantation. Although CKD is a serious disorder, it may curable when diagnosed in the initial stage.

Many researchers have applied different ML techniques to diagnose CKD in the initial stage, but some data preparation techniques are still missing, but these are important for medical data processing. Such as some researchers used the mean-median method to handle missing values, but it’s not ideal for medical data set because the missing value may be more or less than the mean-median value. Besides, Most of the researchers used one or two feature selection techniques but there is a minor correlation among the features. So, it is not an effective way to select the important features by using only one or two techniques. We have kept each important technique and handled all the things according to the need. Our proposed method pre-processes the raw data through a robust data preparation pipeline and passes processed data to the ML classifiers, which increases the learnability of the ML models and enhances the prediction accuracy. In addition, we have proposed a novel hybrid feature selection method to find the most influential attributes. The proposed hybrid feature selection method improves consistency and reduces the dimensionality of the data, which enhances learnability and decreases the learning time of the ML models.

We organize the paper as follows. In Section II, we briefly describe the related works. In section III, the proposed method is written, where the previously mentioned data preparation pipeline and the novel hybrid feature selection method have been detailed. After that in section IV, we have presented the research outcome and discussed the findings. Finally, we conclude the research in section V, including its contribution, limitation, and future work.
\section{Related Works}
In [1], the authors applied the ANN and Random Forest classification algorithms in a CKD data set. They used the Median method for handling missing data and the Chi-squared test technique for feature elimination and selected 20 features out of 24. The experimental result showed that Random Forest provides highest accuracy of  97.12\%.

In another study [3], The researcher suggested a CKD diagnosis approach based on machine learning. They used optimal subset regression and RF for feature extraction. Finally, they applied six ML algorithms and achieved 99.75\% accuracy with Random Forest.

Authors in [4], have worked on a heart disease data set to diagnose the heart disease at the initial stage. There, the Principal Component Analysis (PCA) has been adopted to turn 13 features into only 2 features. Their proposed method can provide 94.06\% clustering accuracy by combining the heuristic k-means clustering with the meta-heuristic Genetic Algorithm.

In [5], the authors used three ML techniques i.e. Logistic Regression, Wide and Deep Learning, and FNN to diagnose CKD. Also, applied the Mean Median method for handling missing data, Min Max Scaler for normalization and SMOTE for oversampling. For both real and over sampled data, FNN produced the best AUC score of 0.99.

In another study [6], to categorize CKD and Not CKD patients, the authors applied SVM using a RBF Kernel. By integrating six criteria, researchers were able to achieve an overall classification accuracy of 94.44\%.

In [7], the authors suggested a kidney disease detection system that uses an artificial classification technique. The top results were determined by analyzing the sensitivity, specificity, and accuracy metrics of SVM. According to the results of the studies, a well developed classifier may obtain a 94.602\% total performance.

The authors in [8] attempted to develop a system that could identify CKD in patients. They presented a novel approach for determining the most relevant features that correlate with CKD, which combines three feature selection strategies (Extra Tree Classifier, RFE, and Uni-variate statistics). The proposed model with less features provides a 97.58\% accuracy.

\section{Proposed Method}
We started by collecting the data set and splitting it into two groups: the training set and the testing set. To prevent data leakage and over-fitting we fit the data preparation techniques only on our training set then apply them to both sets.
In the training set, we encode the categorical features with the One-hot encoder. Then normalized the feature values using the Min-Max Scaler and imputed the missing values with the K-Nearest Neighbor (KNN) Imputation technique. After that, we remove the outliers using the Local Outlier Factor (LOF) method and balance the classes implementing SMOTE. Then we use a novel hybrid feature selection technique to select only important features. After selecting the important predictors, we standardize the values using the Standard Scaler and train the Machine Learning models. Then we validate the models using the Repeated Stratified K-fold Cross-validation. After that, we choose the high-performing models to evaluate the models using the testing set.
In the testing set, we take only important features according to the feature selection step of the training set, then we encode the categorical features using the One-hot encoding technique. Then we apply the Min-Max Normalization, KNN Imputation, and Standardization.  
Finally, we evaluate the selected trained models using the testing data set. Fig. 1 illustrates the workflow of our methodology and the entire works are as follows. 
\begin{figure}[hbt!]
\centerline{\includegraphics[width=260pt]{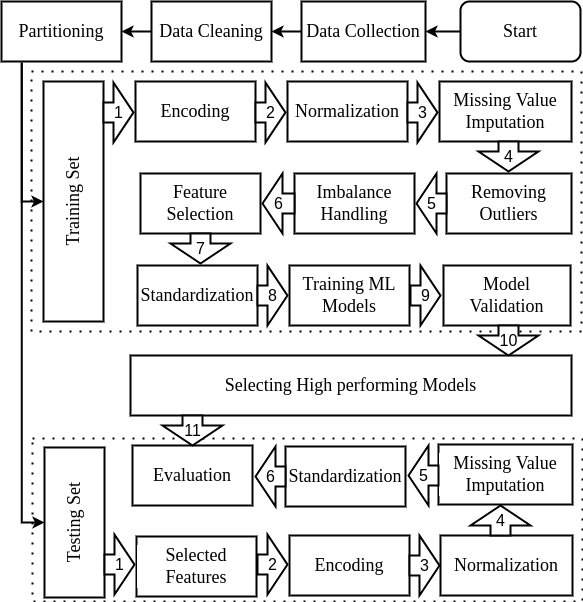}}
\caption{Workflow of the Method}
\label{figure}
\end{figure}

\centerline{Fig. 1. Workflow of the Method}
\subsection{Preliminaries}\label{AA}
The preliminaries to building the models are specified in this section, which describes the data set and the working environment.

\paragraph{Data Collection and Cleaning}
We have collected a data set named Chronic Kidney Disease Data Set from the University of California Irvine Machine Learning Repository [9]. The data set has 400 samples and 24 features, with a lot of missing values, outliers, typing mistakes, class imbalances, etc. In 400 samples, 250 belong to ckd and 150 belong to notckd. The features Packed Cell Volume (pcv), Red Blood Cell Count (rc), and White Blood Cell Count (wc) were mistakenly considered as nominal because of typing errors. We have converted those features into numerical data types. Similarly, few other features also have some typos, which have been fixed by the python map method.

\paragraph{Experimental Setup}
All the techniques have been performed in Python (Version  3.7.12) using Google Collaboratory, including packages or modules such as Scikit Learn (1.0.1), NumPy (1.11.0), Pandas (1.1.5) being used. 
\subsection{Data Partitioning}
The goal of utilizing Machine Learning algorithms to create a model is to simulate real-life data that has never been seen before and to find out how to forecast or categorize it consistently. If data leakage occurs, a model can not perform effectively when face new data in the actual world. While developing a model It is essential to take precautions to avoid data leaking. In the first step to avoiding data leakage, we have checked the duplicate records from the data set but there was not any single duplicate sample, then followed the Cross-Validation and Testing approach for data partitioning, where 75\% of data have been separated for the Training set and 25\% for the Testing set. Different models are trained and validated with the Training set and the best parameters and top-performing models are selected. Finally, we have evaluated the selected models with the Testing set. 

\subsection{Data Preparation}
Different data preparation methods have been prepared based on the Training set only and applied on both partitions. So that the local maxima and local minima for each technique of data preparation are different between the Training and Testing set.

\paragraph{Encoding}
After cleaning the data set, there were 10 categorical (nominal) features. The values of those categorical features have been encoded into 0’s or 1’s with the One-hot Encoding technique. Table I shows which categorical values have been encoded into 1 or 0.
\begin{table}[hbt!]
\centering
\caption{Encoding categorical features using the One-hot encoder}
\label{tab:my-table}
\begin{tabular}{|l|l|l|}
\hline
\multicolumn{1}{|c|}{\textbf{Categorical Feature}} & \multicolumn{1}{c|}{\textbf{1}} & \multicolumn{1}{c|}{\textbf{0}} \\ \hline
Red Blood Cells (rbc)         & normal  & abnormal   \\ \hline
Pus Cell (pc)                 & normal  & abnormal   \\ \hline
Pus Cell Clumps (pcc)         & present & notpresent \\ \hline
Bacteria (ba)                 & present & notpresent \\ \hline
Hypertension (htn)            & yes     & no         \\ \hline
Diabetes Mellitus (dm)        & yes     & no         \\ \hline
Coronary Artery Disease (cad) & yes     & no         \\ \hline
Pedal Edema (pe)              & yes     & no         \\ \hline
Anemia (ane)                  & yes     & no         \\ \hline
Appetite (appet)              & poor    & good       \\ \hline
\end{tabular}
\end{table}
\paragraph{Normalization}
In [10] authors mentioned that normalization improves Machine Learning's ability to detect patterns in data by measuring similarities and distances between samples. Besides, we mentioned earlier that the data set holds a lot of missing values, which have been imputed in the next step using the KNN Imputation technique. As KNN Imputation works through measuring distance, directly imputing based on raw data can lead to the wrong imputation. To avoid this we have rescaled the features using the Min-Max scaler. It rescales features ranges between 0 and 1 using $(1)$ [10]. Where $x$ denotes the real value,  $x_{min}$ and $x_{max}$ are min and max values of the feature $f i$, the rescaled value is $x_n$ and $x, x_{min}, x_{max} \in f_i$. \begin{equation}x_n = \frac{x - x_{min}}{x_{max} - x_{min}}\end{equation}
\paragraph{Missing Value Imputation}
The data set contains 24 features and 400 records, where 1008 values are missing. To handle those missing values, we have used the K-Nearest Neighbour (KNN) Imputation technique. For medical data pre-processing KNN Imputation provides most effective outcome comparing to mean and median [13]. The K-Nearest Neighbours (KNN) recognizes k-complete neighboring points using certain distance metrics, normally euclidean distance, and missing data can be estimated using completed values of neighboring observational data. The mean value of the k-neighbors found in the data set is used to impute the missing values of each sample. We have considered k = 5 and took uniform weight for the KNN Imputation technique. Uniform weights consider all nearest neighbours as equal weighted.

\paragraph{Outlier Removing}
Outliers have a terrible impact on training a Machine Learning model. Because it can lead to training a model from irrelevant data. In [11], the author has written that when effective outlier detection techniques are implemented, Machine Learning models on medical data analysis will deliver more accurate outcomes and diseases may be recognized. According to the data set and problem statement of this research, we have implemented the Local Outlier Factor (LOF) technique, which performs a density-based detection to remove outliers. The density-based outlier detection methods works better than the statistical and distance-based methods [11]. In this technique, the locality is determined by calculating the difference between the k-nearest neighbors and used to calculate the density of the local population. Outliers are extracted from samples by comparing between one sample’s local density and k-nearest neighbors density. We considered 20 nearest neighbours, calculated the Euclidean distance, and found around 26 samples containing outliers. After removing those we get around 274 samples from the training data set.

\paragraph{Data Imbalance Handling}
Class imbalance is a general issue with the real world data set in almost every domain and affects the accuracy of machine learning models. Because the Machine Learning model provides poor accuracy towards the lesser number of data class [12]. The authors of [13] mentioned that when there are a lot of samples in the data set, under-sampling might be a good way to handle data imbalance. But when the data set is imbalanced but small, the Synthetic over-sampling technique works best. In our research, we have applied the Synthetic Minority Oversampling Technique (SMOTE). Because after removing outliers, we get around 274 samples only, where 114 patients had CKD and others 184 were not CKD affected. For the minority class, synthetic samples are prepared in SMOTE. This approach also aids in overcoming the problem of over-fitting caused by random oversampling. Table II shows the frequency of the CKD in various steps of data preparation. Here Yes \& No represent the number of CKD positive \& negative patient respectively.
\begin{table}[hbt!]
\centering
\caption{Data of CKD positive \& negative patients in different steps}
\label{tab:my-table}
\begin{tabular}{|l|cc|cc|cc|}
\hline
             & \multicolumn{2}{c|}{13 Features}   & \multicolumn{2}{c|}{08 Features}   & \multicolumn{2}{c|}{05 Features}   \\ \hline
             & \multicolumn{1}{c|}{Yes} & No & \multicolumn{1}{c|}{Yes} & No & \multicolumn{1}{c|}{Yes} & No \\ \hline
Training Set & \multicolumn{1}{c|}{116} & 184     & \multicolumn{1}{c|}{116} & 184     & \multicolumn{1}{c|}{116} & 184     \\ \hline
LOF          & \multicolumn{1}{c|}{114} & 160     & \multicolumn{1}{c|}{113} & 166     & \multicolumn{1}{c|}{109} & 158     \\ \hline
SMOTE        & \multicolumn{1}{c|}{160} & 160     & \multicolumn{1}{c|}{169} & 169     & \multicolumn{1}{c|}{158} & 158     \\ \hline
\end{tabular}
\end{table}
\paragraph{Proposed Novel Hybrid Feature Selection Method}
In Healthcare Informatics, identifying the most significant risk factors helps to remove redundant attributes, enhance the consistency of data, reduce the training time of ML algorithms, and improve the prediction performance [14]. Since the 1970s, feature selection has been an effectual field of research and development by increasing learning efficiency, improving learning performance such as predictive accuracy, and improving the comprehensibility of learned results [15]. In [4], to decrease features, Principal Component Analysis (PCA) was adopted. In [1], authors extracted the most significant features using the Chi-squared test. Besides, the researchers have been using different hybrid feature selection approaches in recent years because these produce better outcomes than traditional methods [14].

Here we propose and apply a novel hybrid feature selection method. The method aggregates the Wrapper, Filter and Ensemble feature selection methods. Here Chi-Squared Test ($Chi^2$) and Mutual Information (MI) have been applied from the Wrapper Method. Recursive Feature Elimination with Cross-Validation (RFECV) along with multiple estimators have been implemented from the Filter Method. Random Forest and Decision Tree classifiers have been used from the Ensemble Methods. The approach eliminates the redundant features and imparts three subsets $F_{1}, F_{2}, F_{3}$ with important features, where $F_{1}, F_{2}, F_{3} \in Features (F)$. The formula of $F_{1}, F_{2}, F_{3}$ are (2), (3), and (4) respectively.

\begin{equation}F_{1} = \{S_w \cup S_f \cup S_e\} - \{S_{cor}\}\end{equation}\begin{equation}F_{2} = \{(S_w \cap S_f) \cup (S_w \cap S_e) \cup (S_f \cap S_e) \} - \{S_{cor}\}\end{equation}\begin{equation}F_{3} = \{S_w \cap S_f \cap S_e\} - \{S_{cor}\}\end{equation}Equations (2), (3), and (4) are derived from the set operations below.
\begin{itemize}
\item $S_w = \{ x | x \in Chi^2 \cap MI\}$\\
\item $S_f = \{x | x \in (R_G \cap  R_R) \cup (R_G \cap R_L) \cup (R_R \cap R_L)\}$\\
\item $S_e = \{x|x \in RF \cap DT\}$\\
\item $S_{cor}$ = $\{x|x \in$ Highly correlative features$\}$\\
\item $Chi^2$ = $\{i|i \in$ Top 70\% important features found from the Chi-Squared Test$\}$\\
\item $MI = \{i|i \in$ Top 70\% important features found from the Mutual Information technique$\}$\\
\item $R_G$ = $\{i|i \in$ Important features found from the RFECV and Gradient Boosting estimator$\}$\\
\item $R_R$ = $\{i|i \in$ Important features found from the RFECV and Random Forest estimator$\}$\\
\item $R_L$ = $\{i|i \in$ Important features found from the RFECV and Logistic Regression estimator$\}$\\
\item $RF$ = $\{i|i \in$ Important features found from the Random Forest Algorithm$\}$\\
\item $DT$ = $\{i|i \in$ Important features found from the Decision Tree Algorithm$\}$
\end{itemize}

Equations (1), (2), (3) impart feature set $F_1$, $F_2$, $F_3$ which contain 13, 08, and 05 number of important features respectively. Below is the elaboration of the mentioned hybrid feature elimination approach for our study. After applying the Chi-Squared test (Chi2) and Mutual Information (MI), we select only the top 70\% features for further analysis and hold these features into two unique sets $Chi^2$ and $MI$ respectively. We considered f-values for the Chi-Squared test and 03 nearest neighbors for Mutual Information.
Then we apply the RFECV. To find the most important attributes, we take three different estimators and 10 fold cross validation on the RFECV. Those estimators are the Gradient Boosting classifier with 1000 boosting stages and rate of learning 0.01, the Random Forest classifier with 1000 trees in the forest, and the Logistic Regression with LBFGS solver along with maximum iteration 100. Then we hold the important features found from these three different estimators into three unique sets $R_G, R_R, R_L$ respectively.
Finally we apply two ensemble algorithms and validate 5 times to find most important features. There the ensemble algorithms are the Random Forest Classifier with 1000 trees  and the Decision Tree Classifier with the Entropy criterion for information gain. Then  hold these important features in the RF and DT set respectively.
Now, The hybrid feature selection technique creates 3 initial sets $S_w, S_f, S_e$ from above sets $Chi^2, MI, R_G, R_R, R_L, RF$, and $DT$ using formulas mentioned earlier.

To reduce redundancy identifying correlative features is also important. We conduct the Pearson  Correlation Check and find only 07 features have more than 70\% correlation. Fig. 2 depicts the correlation among features and Table III shows the correlative features (more than 70\% correlation) found from the Pearson Correlation Check. To eliminate features with highly correlated correlation, we have only considered those features which have over 85\% correlation and hold the highly correlative feature in set $S_{cor}$. Here only Serum Creatinine (sc) and Blood Urea (bu) are the highly correlative features and we hold the Blood Urea into the set $S_{cor}$. 

\begin{figure}[hbt!]
\centerline{\includegraphics[width=245pt]{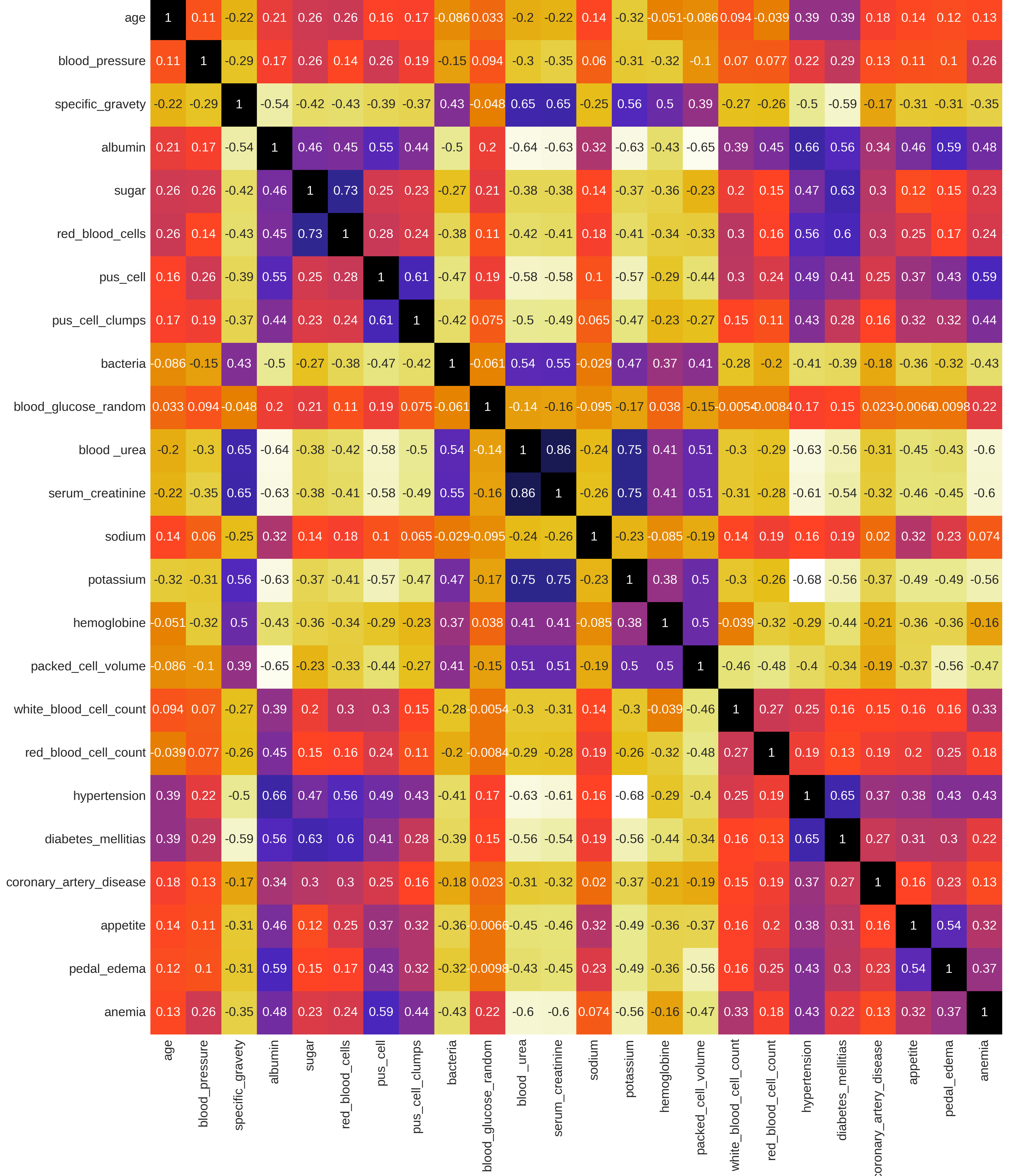}}
\label{figure}
\centerline{Fig. 2. Heat map of feature correlation}
\end{figure}

\begin{table}[hbt!]
\caption{Correlative features over 70 \% correlation}
\label{tab:my-table}
\begin{tabular}{|l|c|}
\hline
\multicolumn{1}{|c|}{\textbf{Correlative Features}} & \textbf{Percentage of Correlation} \\ \hline
Red Blood Cells , Sugar         & 74 \\ \hline
Serum Creatinine, Blood Urea    & 87 \\ \hline
Potassium, Blood Urea           & 78 \\ \hline
Potassium, Serum Creatinine     & 78 \\ \hline
Diabetes Mellitus, Hypertension & 71 \\ \hline
\end{tabular}
\end{table}
 
Finally, the hybrid feature selection technique imparts three subsets $F_{1}, F_{2}, F_{3}$ using (2), (3), (4). Here $F_1 = $ $\{$Specific Gravity, Albumin, Sugar, Serum Creatinine, Hemoglobin, Packed Cell Volume, Red Blood Cells, Pus Cell, Pus Cell Clumps, Hypertension, Diabetes Mellitus, Appetite, and Pedal Edema$\}$, $F_2 = $ $\{$ Specific Gravity, Albumin, Serum Creatinine, Hemoglobin, Red Blood Cells, Pus Cell Clumps, Hypertension, and Diabetes Mellitus$\}$, and $F_3 = $ $\{$Specific Gravity, Serum Creatinine, Hemoglobin, Pus Cell Clumps, and Hypertension$\}$.
We have further used these 3 different subsets of important features separately to train the distinct machine learning classifiers.

\paragraph{Standardization}
Authors in [5] mentioned that data standardization has a substantial impact on the Machine Learning algorithm's performance. Model learning and classification performance can be improved by standardizing features. Before we start building predictive models, we need to do one last step. We standardized data using the Standard Scaler from the Scikit-learn [16] machine learning library  to ensure that all features are measured at the same level. The standard value of sample $x$ is calculated from (5). Where $u$ is the training sample's mean and $s$ is the training sample's standard deviation. \begin{equation}z = \frac{(x - u)}{s}\end{equation}
\subsection{Applying Machine Learning Algorithms}
The following Machine Learning models for diagnosing CKD were developed by applying ML algorithms on the distinct set of features found in the novel hybrid feature selection technique. 
\begin{enumerate}
\item Support Vector Machine (SVM)
\item Gaussian Naive Bayes (Gaussian NB)
\item Decision Tree (DT)
\item Random Forest (RF)
\item Logistic Regression (LR)
\item K-Nearest Neighbor (KNN)
\item Gradient Boosting (GBoost)
\item Adaptive Boosting (AdaBoost)
\item Extreme Gradient Boosting (XGBoost)
\end{enumerate}

In the SVM classifier we have used the RBF kernel, where parameter of regularisation $c$ is 1.0 and gamma is found from  (6). \begin{equation}\gamma =  \frac{1}{n\ features\ *\ \sigma^2}\end{equation}For the Decision Tree classifier the Gini impurity has been used and choose the best split at each node. 1000 estimators have been used for the Random Forest. The Logistic Regression (LR) model is trained using the LBFGS solver. The KNN is trained with 25 nearest neighbor, where the distance metric is Minkowski with p-value 2. In training the Gradient Boosting algorithm, 1000 boosting stages have been used as the estimator, where learning rate = 0.01, to assess a split's efficiency we have used the Mean Squared Error (MSE) with improvement score by Friedman and the loss function has been optimized using Logistic Regression. For the Adaptive Boosting algorithm, the range of estimators that can be used is limited to 50, SAMME.R is the ground estimator, and the learning rate is 1.0. Lastly, 1000 estimator have been used to train the Extreme Gradient Boosting algorithm.
\subsection{Model Validation}
Validating the above models, we have used the Repeated Stratified K-fold Cross-validation. It keeps the original data set's class ratio throughout the K folds and repeats n times. When a model is trained, stratified k-fold cross-validation prevents data leakage and over-fitting [16]. We have applied the 10-fold Cross-validation repeatedly 10 times using the Repeated Stratified K-fold Cross-validation.  The Table IV shows the mean accuracy of the models which are trained with different numbers of features found in (1), (2), and (3). The equations (1), (2), (3) impart feature set $F_1$, $F_2$, $F_3$ which contain 13, 08, and 05 important features respectively.

\begin{table}[hbt!]
\begin{center}
\caption{Model validation with 10 stratified 10-fold cross-validation}
\label{tab:my-table}
\begin{tabular}{|l|c|c|c|}
\hline
\multicolumn{1}{|c|}{\textbf{}}      & \multicolumn{3}{c|}{\textbf{Mean Accuracy}}                        \\ \hline
\multicolumn{1}{|c|}{\textbf{Model}} & \textbf{13 Features} & \textbf{08 Features} & \textbf{05 Features} \\ \hline
SVM        & \textbf{100.00} & 99.39 & 98.34 \\ \hline
GaussianNB & \textbf{100.00}  & 99.14 & 88.23 \\ \hline
DT         & 99.63  & 98.93 & 97.95 \\ \hline
RF         & \textbf{100.00} & \textbf{100.00} & \textbf{98.91} \\ \hline
LR         & \textbf{100.00}  & 99.69 & 98.34 \\ \hline
KNN        & 98.33  & 99.51 & 97.81 \\ \hline
GBoost     & 99.50  & 99.05 & 98.31 \\ \hline
AdaBoost   & 99.82  & 99.89 & 98.61 \\ \hline
XGBoost    & \textbf{100.00} & 99.51 & 98.18 \\ \hline
\end{tabular}
\end{center}
\end{table}
From the beginning of the method, we concentrated on each aspect that can affect the learning of a Machine Learning classifier and we get satisfactory results with most of the models. After validation, the ML models trained with the set of 13 features show the best result. However, the remaining ML models also provide excellent outcomes. In the next section, we have tested the trained models with the testing data set and showed the evaluation outcomes.
\section{Result \& Discussion}
After collecting the data set, we have shuffled and separated it into training and testing set. To avoid data leakage and over-fitting, various data preparation techniques have been fitted only on the training set, then applied to both training and testing sets. In the feature selection step, we have applied a novel hybrid feature selection method that returns 03 subsets of features. Then trained the ML models using those distinct subsets. We have considered and evaluated the ML models with every subset of important features because of their excellent performance in the model validation. Table V, VI and VII show the evaluation results of the trained models after evaluating with the processed testing data set according to fig. 1. We have used the Accuracy, F1 - Score, and AUC Score matrices to assess model effectiveness. The equations (7), (8), (9) [16][18] below represents the formula of Accuracy, F1 - Score, and AUC Score respectively.\begin{equation}
Accuracy = \frac{TP + TN}{TP + TN + FP + FN}
\end{equation}
\begin{equation}
F1-score = 2 * \frac{PPV * TPR}{PPV + TPR}
\end{equation}\begin{equation}
AUC(f) = \frac{\sum_{t_0 \in D^0} * \sum_{t_0 \in D^1 * 1[f(t_0) < f(t_1)]}}{|D^0| * |D^1|}
\end{equation}\[Where\ TP = True\ Positive\] \[TN = True\ Negative\] \[FP = False\ Positive\] \[FN = False\ Negative\] \[TPR\ or\ Recall = \frac{TP}{TP + FN}\]\[PPV or Precision = \frac{TP}{TP + FP}\] \[1[f(t_0) < f(t_1)] = 1, while\ f(t_0) < f(t_1)\ is\ True\ otherwise\ 0\] \[D^0 = Set\ of\ negative\ example\] \[D^1 = Set\ of\ positive\ example\]

Table V shows the evaluation outcome of different ML models using feature set $F_1$ found in (2). The $F_1$ set contains 13 important features to train and evaluate the models which are Specific Gravity, Albumin, Sugar, Serum Creatinine, Hemoglobin, Packed Cell Volume, Red Blood Cells, Pus Cell, Pus Cell Clumps, Hypertension, Diabetes Mellitus, Appetite, and Pedal Edema. Here, the Random Forest model provides 100\% accuracy and the lowest accurate model is K-Nearest Neighbor (KNN) because of 95\% accuracy.
\begin{table}[hbt!]
\begin{center}
\caption{Model Evaluation with 13 features}
\label{tab:my-table}
\begin{tabular}{|l|c|c|c|}
\hline
\multicolumn{1}{|c|}{\textbf{Model}} &
  \textbf{Accuracy} &
  \textbf{\begin{tabular}[c]{@{}c@{}}F1-Score\\ (macro avg.)\end{tabular}} &
  \textbf{\begin{tabular}[c]{@{}c@{}}AUC\\ (macro avg.)\end{tabular}} \\ \hline
SVM         & 98  & 98  & 98.48 \\ \hline
Gaussian NB & 99  & 99  & 98.52 \\ \hline
DT          & 99  & 99  & 99.24 \\ \hline
RF          & \textbf{100} & \textbf{100} & \textbf{100.00}   \\ \hline
LR          & 97  & 97  & 97.73    \\ \hline
KNN         & 95  & 95  & 96.21 \\ \hline
GBoost      & 99  & 99  & 99.24 \\ \hline
AdaBoost    & 97  & 97  & 97.72 \\ \hline
XGBoost     & 99  & 99  & 99.24 \\ \hline
\end{tabular}
\end{center}
\end{table}

Table VI shows the evaluation outcome of different ML models using feature set $F_2$ found in (3). The $F_2$ set contains 08 important features to train and evaluate the models which are Specific Gravity, Albumin, Serum
\begin{table}[hbt!]
\begin{center}
\caption{Model Evaluation with 08 features}
\label{tab:my-table}
\begin{tabular}{|l|c|c|c|}
\hline
\multicolumn{1}{|c|}{\textbf{Model}} &
  \textbf{Accuracy} &
  \textbf{\begin{tabular}[c]{@{}c@{}}F1-Score\\ (macro avg.)\end{tabular}} &
  \textbf{\begin{tabular}[c]{@{}c@{}}AUC\\ (macro avg.)\end{tabular}} \\ \hline
SVM         & 99  & 99  & 98.53 \\ \hline
Gaussian NB & 99 & 99 & 99.24   \\ \hline
DT          & 98  & 98  & 98.49 \\ \hline
RF          & \textbf{100} & \textbf{100} & \textbf{100.00}   \\ \hline
LR          & 99  & 99  & 99.24 \\ \hline
KNN         & 97  & 97  & 97.73 \\ \hline
GBoost      & 98  & 98  & 98.48 \\ \hline
AdaBoost    & 98  & 98  & 98.48 \\ \hline
XGBoost     & 98  & 98  & 98.48 \\ \hline
\end{tabular}
\end{center}
\end{table}Creatinine, Hemoglobin, Red Blood Cells, Pus Cell Clumps, Hypertension, and Diabetes mellitus. Finally, the Random Forest (RF) model provide 100\% accuracy. On the other hand, the K-Nearest Neighbor (KNN) model provides the lowest accuracy of 97\%.

Table VII shows the evaluation outcome of different ML models using feature set $F_3$ found in (4). The $F_3$ set contains 08 important features to train and evaluate the models which are Specific Gravity, Serum Creatinine, Hemoglobin, Red Blood Cells, and Hypertension. There the Support Vector Machine (SVM), Random Forest (RF), Gradient Boosting (GBoost), and Extreme Gradient Boosting (XGBoost) model provide the highest accuracy of 99\% but the SVM model performs best because of the highest AUC Score (99.24\%). Whereas the K-Nearest Neighbor (KNN) model has the lowest accuracy, with a score of 97\%.

\begin{table}[hbt!]
\begin{center}
\caption{Model Evaluation with 05 features}
\label{tab:my-table}
\begin{tabular}{|l|c|c|c|}
\hline
\multicolumn{1}{|c|}{\textbf{Model}} &
  \textbf{Accuracy} &
  \textbf{\begin{tabular}[c]{@{}c@{}}F1-Score\\ (macro avg.)\end{tabular}} &
  \textbf{\begin{tabular}[c]{@{}c@{}}AUC\\ (macro avg.)\end{tabular}} \\ \hline
SVM         & \textbf{98} & \textbf{98} & 97.77 \\ \hline
Gaussian NB & 82 & 82 & 86.36 \\ \hline
DT          & 96 & 96 & 96.30 \\ \hline
RF          & 97 & 97 & 97.01 \\ \hline
LR          & 95 & 95 & 96.21 \\ \hline
KNN         & \textbf{98} & \textbf{98} &\textbf{ 98.48} \\ \hline
GBoost      & \textbf{98} & \textbf{98} & 97.78 \\ \hline
AdaBoost    & \textbf{98} & \textbf{98} & 97.78 \\ \hline
XGBoost     & 97 & 97 & 96.30 \\ \hline
\end{tabular}
\end{center}
\end{table}

Now, we critically analyze the findings from Tables V, VI, and VII. Fig. 3 illustrates the analytical comparison among these tables and Table VIII shows the accuracy comparison. The analysis shows that the SVM achieves the highest Accuracy of 99\% and AUC of 98.53\% when trained with $F_2$, Gaussian NB gains the highest Accuracy of 99\% with both $F_1$ and $F_2$ but achieves the highest AUC score of 98.52\% with only $F_1$. The DT performs best (99\% Accuracy and 99.24\% AUC) with $F_1$. The LR reaches 99\% and 99.24\% of Accuracy and AUC score respectively with $F_2$. The KNN attains 98\% Accuracy with $F_3$ but gains the highest AUC of 99.21\% with $F_1$. The GBoost
\begin{table}[hbt!]
\centering
\caption{Accuracy comparison of Table V, VI and VII}
\label{tab:my-table}
\begin{tabular}{|c|ccc|}
\hline
            & \multicolumn{3}{c|}{\textbf{Accuracy}}                                              \\ \hline
\textbf{Models} & \multicolumn{1}{c|}{\textbf{13 Feature}} & \multicolumn{1}{c|}{\textbf{08 Feature}} & \textbf{05 Features} \\ \hline
SVM         & \multicolumn{1}{c|}{98}           & \multicolumn{1}{c|}{\textbf{99}}  & 98          \\ \hline
Gaussian NB & \multicolumn{1}{c|}{\textbf{99}}  & \multicolumn{1}{c|}{\textbf{99}}  & 82          \\ \hline
DT          & \multicolumn{1}{c|}{\textbf{99}}  & \multicolumn{1}{c|}{98}           & 97          \\ \hline
RF          & \multicolumn{1}{c|}{\textbf{100}} & \multicolumn{1}{c|}{\textbf{100}} & 97          \\ \hline
LR          & \multicolumn{1}{c|}{97}           & \multicolumn{1}{c|}{\textbf{99}}  & 95          \\ \hline
KNN         & \multicolumn{1}{c|}{95}           & \multicolumn{1}{c|}{97}           & \textbf{98} \\ \hline
GBoost      & \multicolumn{1}{c|}{\textbf{99}}  & \multicolumn{1}{c|}{98}           & 98          \\ \hline
AdaBoost    & \multicolumn{1}{c|}{97}           & \multicolumn{1}{c|}{\textbf{98}}  & \textbf{98} \\ \hline
XGBoost     & \multicolumn{1}{c|}{\textbf{99}}  & \multicolumn{1}{c|}{98}           & 97          \\ \hline
\end{tabular}
\end{table}
and XGBoost both model achieves the highest accuracy of 99\% and AUC of 99.24\% when trained with $F_1$, AdaBoost achieves the highest accuracy of 98\% with both $F_2$  and $F_3$ whereas achieving the highest AUC of 99.72\% with $F_1$. Last, the RF model achieves 100\% of Accuracy and AUC with both $F_1$ and $F_2$. 

\begin{figure}[hbt!]
\centerline{\includegraphics[width=255pt]{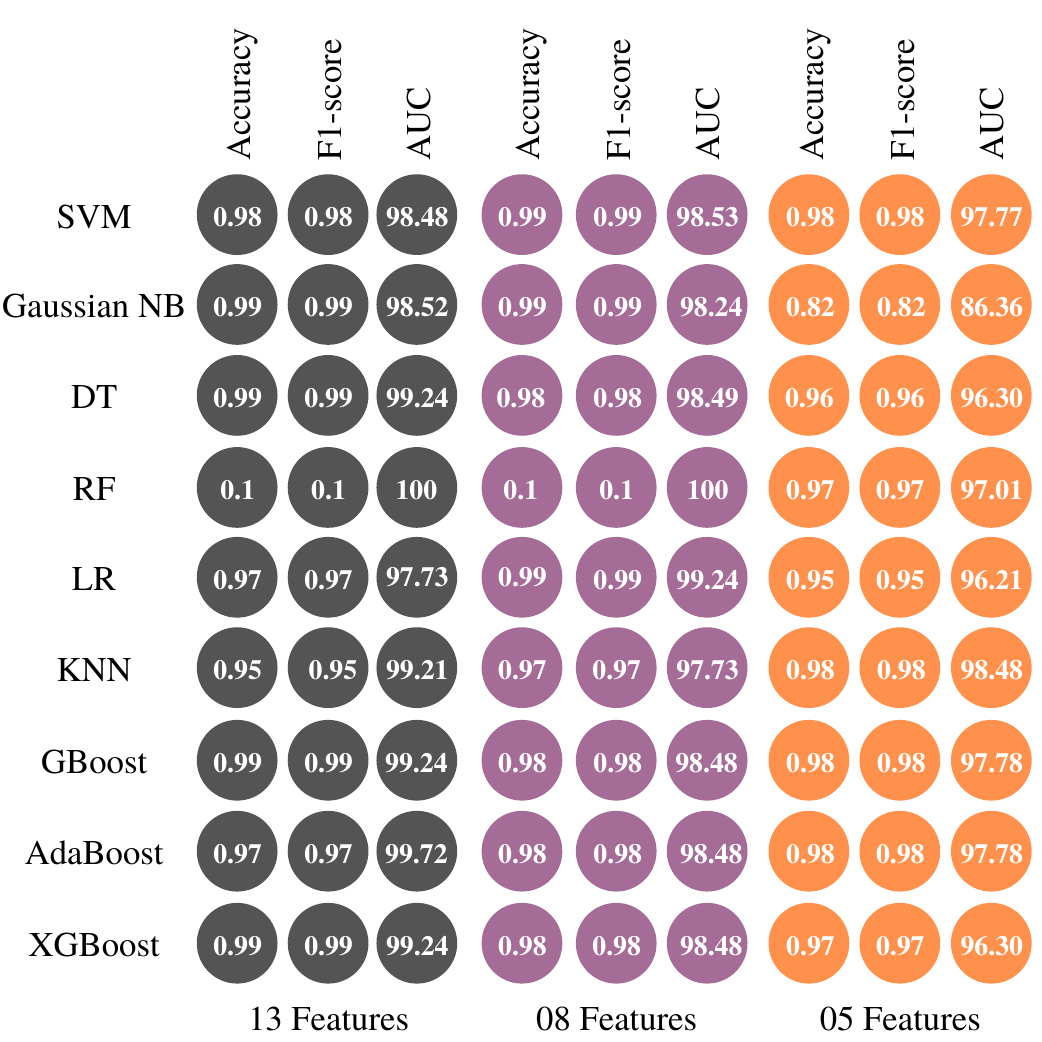}}
\caption{Result comparison}
\label{figure}
\centerline{Fig. 3. Result comparison of Table V, VI and VII}
\end{figure}

Considering Fig. 3 and Table VIII, we can come to the decision that the ML models perform comparatively higher in terms of AUC score when trained using the feature set $F_1$ and achieve a higher Accuracy score when trained using set $F_2$. From the perspective of the medical data set, the AUC score is more meaningful compared to the Accuracy score. So we ultimately select those ML models which are trained using the feature set $F_1$. Fig. 4 illustrates the Accuracy and AUC score of our finally selected models.
\begin{figure}[hbt!]
\centerline{\includegraphics[width=255pt]{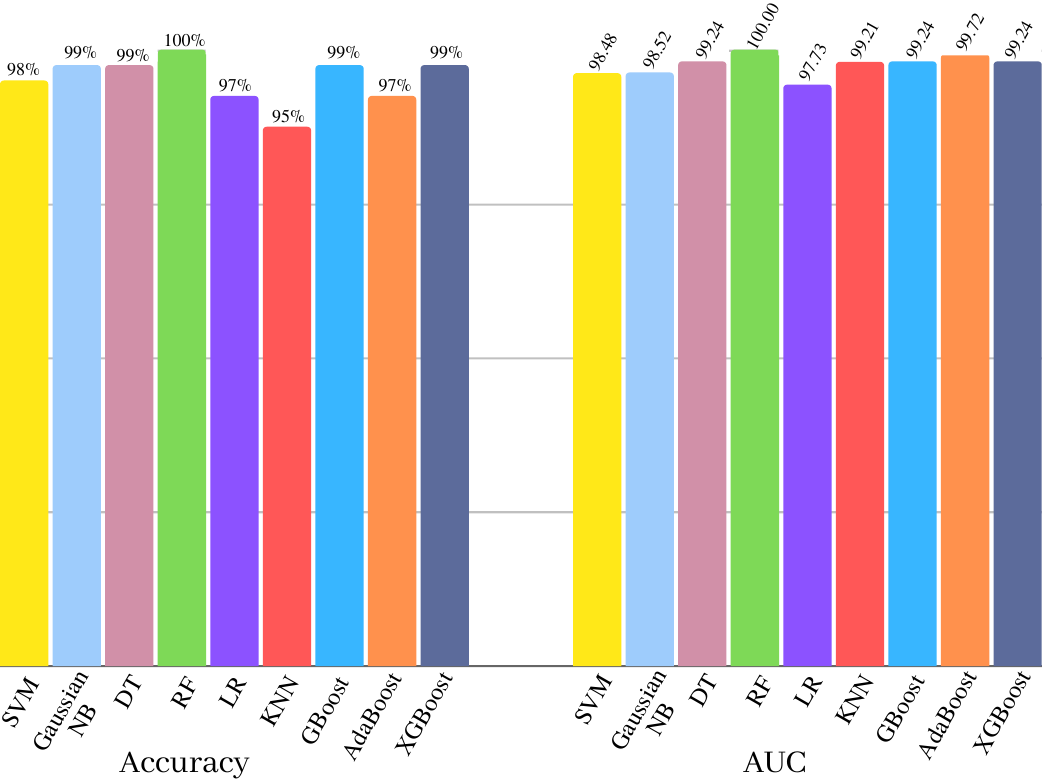}}
\caption{Performance of selected ML models using 13 features}
\label{figure}
\centerline{Fig. 4. Performance of selected ML models using 13 features}
\end{figure}Here Fig. 4 depicts that the SVM, Gaussian NB, DT, RF, LR, KNN, GBoost, AdaBoost, XGBoost respectively achieve accuracy of 98\%, 99\%, 99\%, 100\%, 97\%, 95\%, 99\%, 97\%, 99\% and AUC of 98.48\%, 98.52\%, 99.24\%, 100\%, 97.73\%, 99.21\%, 99.24\%, 99.72\%, 99.24\%.

In Table IX, we compare the accuracy of our method with some related works. The comparative study shows that the proposed method provides higher accuracy in detecting CKD at the initial stage than the previous works.
\begin{table}[hbt!]
\centering
\caption{Comparative analysis with related works}
\label{tab:my-table}
\begin{tabular}{|ccc|}
\hline
\textbf{Algorithms} &
  \textbf{Accuracy} &
  \textbf{AUC} \\ \hline
\begin{tabular}[c]{@{}c@{}}Random Forest \\ Artificial Neural Network {[}1{]}\end{tabular} &
  \begin{tabular}[c]{@{}c@{}}97.12\%\\ 94.05\%\end{tabular} &
  N/A \\ \hline
  \begin{tabular}[c]{@{}c@{}}Logistic Regression\\ Random Forest\\ Integrated Model {[}3{]}\end{tabular} &
  \begin{tabular}[c]{@{}c@{}}\textbf{98.95}\%\\ 99.75\%\\ 99.83\%\end{tabular} &
  N/A \\ \hline
XG Boost {[}8{]} &
  97.60\% &
  N/A \\ \hline
\begin{tabular}[c]{@{}c@{}}Support Vector Machine\\ Logistic Regression\\ MultiLayer Perceptron {[}10{]}\end{tabular} &
  \begin{tabular}[c]{@{}c@{}}95.00\%\\ 98.10\%\\ 98.10\%\end{tabular} &
  \begin{tabular}[c]{@{}c@{}}97.30\\ \textbf{99.40}\\ 99.50\end{tabular} \\ \hline
\begin{tabular}[c]{@{}c@{}}Random Forest\\ Gradient Boosting \\ XG Boost\\ Logistic Regression\\ Support Vector Machine {[}19{]}\end{tabular} &
  \begin{tabular}[c]{@{}c@{}}99.00\%\\ 98.25\%\\ 98.50\%\\ 97.75\%\\ 97.25\%\end{tabular} &
  \begin{tabular}[c]{@{}c@{}}98.77\\ 98.10\\ 98.50\\ 97.41\\ 97.22\end{tabular} \\ \hline
\begin{tabular}[c]{@{}c@{}}XG Boost\\ Logistic Regression {[}20{]}\end{tabular} &
  \begin{tabular}[c]{@{}c@{}}91.32\%\\ 83.92\%\end{tabular} &
  \begin{tabular}[c]{@{}c@{}}92.60\\ 78.40\end{tabular} \\ \hline
\begin{tabular}[c]{@{}c@{}}Logistic Regression\\ Naive Bayes\\ Artificial Neural Network\\ Random Forest\\ Integrated Model {[}21{]}\end{tabular} &
  \begin{tabular}[c]{@{}c@{}}97.70\%\\ 94.25\%\\ 100.00\%\\ 98.85\%\\ 100.00\%\end{tabular} &
  N/A \\ \hline
\multicolumn{3}{|c|}{\textit{The proposed method of this paper}}\\
\begin{tabular}[c]{@{}c@{}}Random Forest\\ Support Vector Machine\\ Gaussian Naive Bayes\\ Logistic Regression\\ Decision Tree\\ Gradient Boosting\\ XG Boosting\\ Adaptive Boosting\\ K-Nearest Neighbour\end{tabular} &
  \begin{tabular}[c]{@{}c@{}}\textbf{100.00\%}\\ \textbf{98.00\%}\\ \textbf{99.00\%}\\ 97.00\%\\ \textbf{99.00\%}\\ \textbf{99.00\%}\\ \textbf{99.00\%}\\\textbf{ 97.00\%}\\ \textbf{95.00\%}\end{tabular} &
  \begin{tabular}[c]{@{}c@{}}\textbf{100.00}\\\textbf{ 98.48}\\ \textbf{98.52}\\ 97.73\\ \textbf{99.24}\\ \textbf{99.24}\\ \textbf{99.24}\\ \textbf{99.72}\\ \textbf{99.21}\end{tabular} \\ \hline
\end{tabular}
\end{table}
\section{Conclusion}
The principal aim of this research is to deal with the complexities of medical data and find an optimal solution for detecting CKD in its initial stage using ML. We also developed a lighter Machine Learning model using only 13 important features supported by our novel hybrid feature selection technique, which may reduce the time and cost of diagnosis. The research can be extended further by using the real-time data set and adding the domain knowledge in detecting outliers, which will enhance the robustness of the proposed method. 


\vspace{12pt}
\color{red}

\end{document}